\documentclass[10pt,twocolumn,letterpaper]{article}

\usepackage[pagenumbers]{wacv} 

\usepackage{graphicx}
\usepackage{amsmath}
\usepackage{amssymb}
\usepackage{booktabs}
\usepackage{multirow}
\usepackage{algorithm}
\usepackage{dsfont}
\usepackage{tikz} 
\usetikzlibrary{arrows}
\usepackage{bbm}
\usepackage{svg}
\usepackage{multirow}
\usepackage[accsupp]{axessibility}

\usepackage[pagebackref,breaklinks,colorlinks]{hyperref}

\usepackage[capitalize]{cleveref}
\crefname{section}{Sec.}{Secs.}
\Crefname{section}{Section}{Sections}
\Crefname{table}{Table}{Tables}
\crefname{table}{Tab.}{Tabs.}

\def\wacvPaperID{1502} 
\def\confName{WACV}
\def\confYear{2024}

\begin{document}

\title{Cross-feature Contrastive Loss for Decentralized Deep Learning on Heterogeneous Data}

\author{Sai Aparna Aketi\\
Purdue University\\
USA\\
{\tt\small saketi@purdue.edu}
\and
Kaushik Roy\\
Purdue University\\
USA\\
{\tt\small kaushik@purdue.edu}
}
\maketitle

\begin{abstract}
The current state-of-the-art decentralized learning algorithms mostly assume the data distribution to be Independent and Identically Distributed (IID). However, in practical scenarios, the distributed datasets can have significantly heterogeneous data distributions across the agents. In this work, we present a novel approach for decentralized learning on heterogeneous data, where data-free knowledge distillation through contrastive loss on cross-features is utilized to improve performance. Cross-features for a pair of neighboring agents are the features (i.e., last hidden layer activations) obtained from the data of an agent with respect to the model parameters of the other agent. We demonstrate the effectiveness of the proposed technique through an exhaustive set of experiments on various Computer Vision datasets (CIFAR-10, CIFAR-100, Fashion MNIST, Imagenette, and ImageNet), model architectures, and network topologies. Our experiments show that the proposed method achieves superior performance ($0.2-4\%$ improvement in test accuracy) compared to other existing techniques for decentralized learning on heterogeneous data.

\end{abstract}

\section{Introduction}
\label{sec:introduction}

Every day, substantial volumes of data are generated across the globe, offering the potential to train powerful deep-learning models.
Compiling such data for centralized processing is impractical due to communication constraints and privacy concerns.
To address this issue, a new interest in developing distributed learning algorithms \cite{agarwal2011distributed} has emerged. Federated learning (FL) or centralized distributed learning \cite{federated} is a popular setting in the distributed machine learning paradigm. In this setting, the training data is kept locally at the edge devices and a global shared model is learned by aggregating the locally computed updates through a coordinating central server. 
Such a setup requires frequent communication with a central server. This becomes a potential bottleneck \cite{haghighat2020applications} and has led to advancements in decentralized machine learning.

Decentralized learning, a subset of distributed optimization, focuses on learning from data distributed across multiple agents without the need for a central server.
It offers many advantages over the traditional centralized approach in core aspects such as data privacy, fault tolerance, and scalability \cite{nedic2020distributed}.
Research has shown that decentralized learning algorithms can perform comparable to centralized algorithms on benchmark vision datasets \cite{d-psgd}.
One of the key assumptions to achieve state-of-the-art performance by the decentralized algorithms is that the data is independently and identically distributed (IID) across the agents. 
In particular, the data is assumed to be distributed in a uniform and random manner across the agents. 
This assumption does not hold in most real-world settings where the data distributions across the agents are significantly different (non-IID/heterogeneous) \cite{skewscout}. 

The effect of heterogeneous data in a peer-to-peer decentralized setup is a relatively under-studied problem and an active area of research. 
Note that, we mainly focus on a common type of non-IID data, widely used in prior works \cite{qgm, skewscout}: a skewed distribution of data labels across agents.
Recently, several methods were proposed to bridge the performance gap between IID and non-IID data for a decentralized setup. Most of these works either make algorithmic changes to track global information \cite{gt, mt, gut, qgm, relaysgd} or utilize extra communication rounds \cite{ngc, CGA} to obtain second-order gradient information. 
In this work, we explore an orthogonal direction of using a data-free knowledge distillation approach to handle heterogeneous data in decentralized setups.

\begin{figure*}[ht]
\centering
\begin{subfigure}{0.32\textwidth}
    \includegraphics[width=\textwidth]{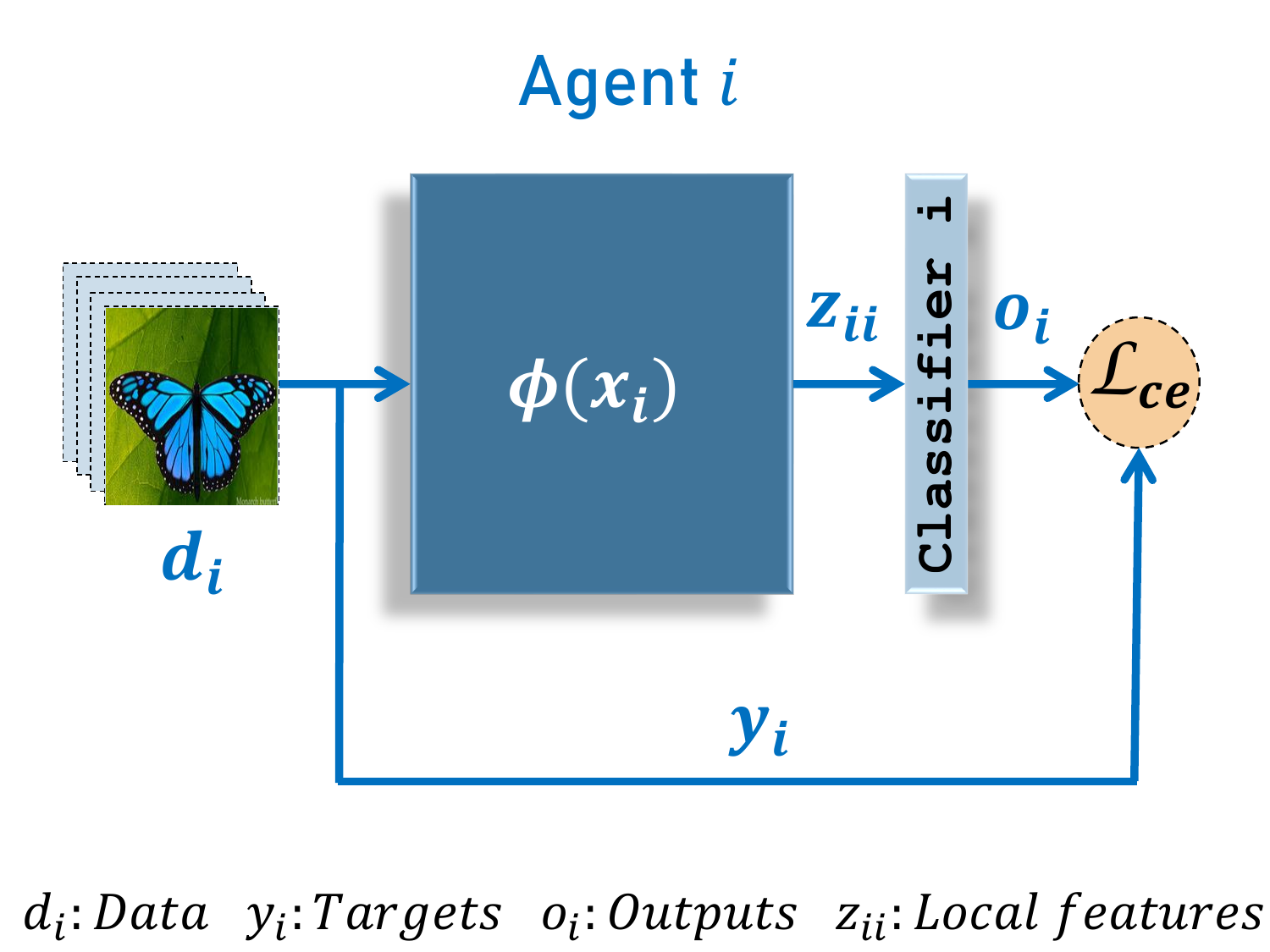}
    \caption{Cross-Entropy Loss ($\mathcal{L}_{ce}$)}
    \label{fig:ce}
\end{subfigure}
\hfill
\begin{subfigure}{0.32\textwidth}
    \includegraphics[width=\textwidth]{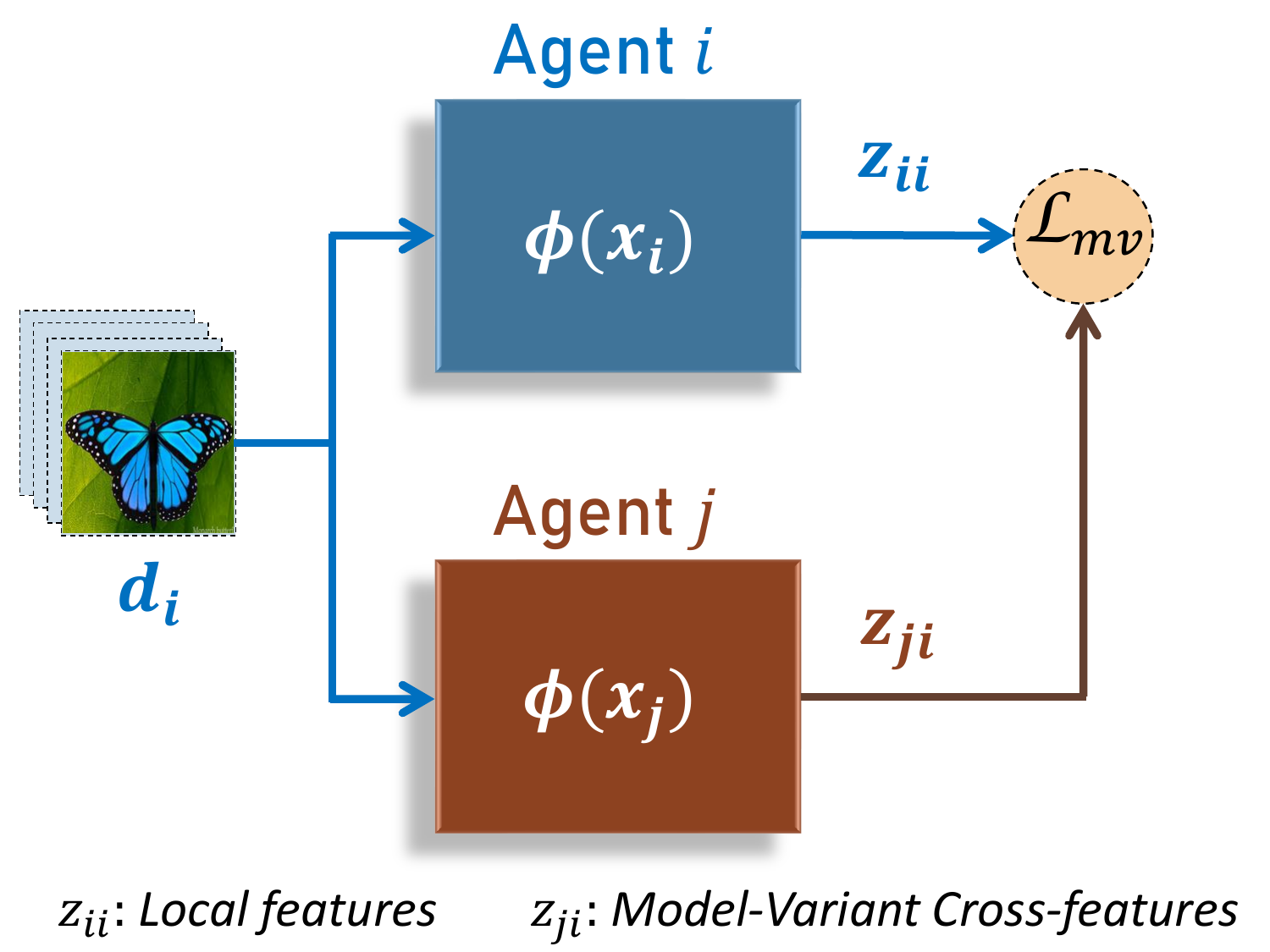}
    \caption{Model-Variant Contrastive Loss ($\mathcal{L}_{mv}$)}
    \label{fig:mvcl}
\end{subfigure}
\hfill
\begin{subfigure}{0.32\textwidth}
    \includegraphics[width=\textwidth]{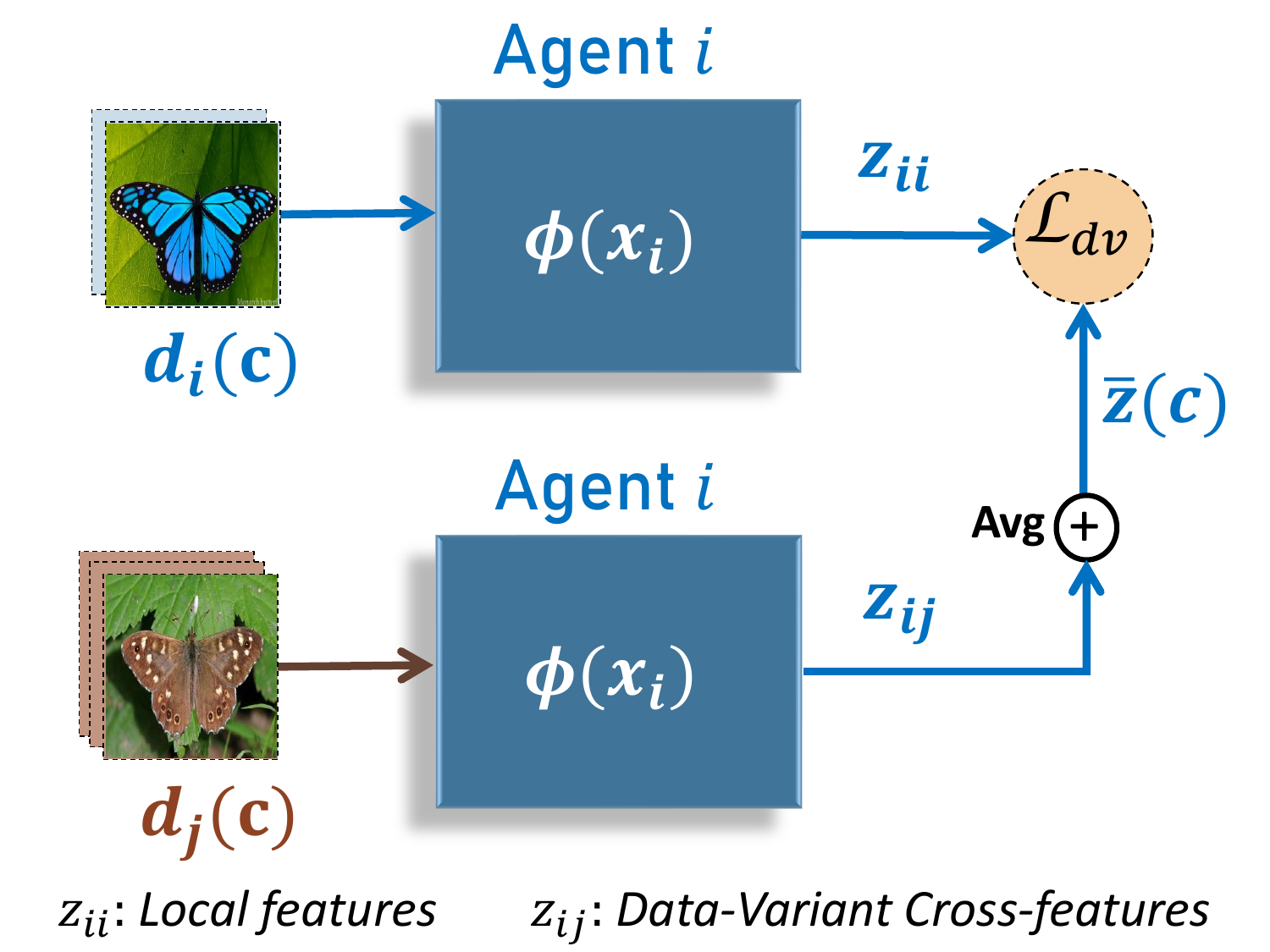}
    \caption{Data-Variant Contrastive Loss ($\mathcal{L}_{dv}$)}
    \label{fig:dvcl}
\end{subfigure}        
\caption{Illustrating different loss components used in the proposed \textit{Cross-features Contrastive Loss}. This illustration describes the loss components with respect to agent $i$ and assumes that it has only one neighbor $j$. The Data-variant Contrastive Loss is only shown for a particular class $c$ and the same rule will be applied to all the classes in parallel. 
$\bar{z}$ is computed at agent $j$ and then communicated to agent $i$.
}
\label{fig:overview}
\end{figure*}

Knowledge distillation methods have been well explored in federated learning (FL) setups with a central server for heterogeneous data \cite{li2019fedmd, lin2020ensemble, zhu2021data, scalable, idkd}. 
However, these approaches leverage the central server and/or need public dataset access and thus, are not transferable to decentralized setups. In this paper, we propose \emph{Cross-feature Contrastive Loss (CCL)} that improves the performance of decentralized training on heterogeneous data when used along with cross-entropy loss ($\mathcal{L}_{ce}$) at each agent. 
In particular, at each agent, we introduce two additional contrastive loss terms on cross-features - (a) model-variant contrastive loss ($\mathcal{L}_{mv}$) and (b) data-variant contrastive loss ($\mathcal{L}_{dv}$).
Cross-features for a pair of neighboring agents are the features (i.e., last hidden layer activations) obtained from the data of an agent with respect to the model parameters of the other agent. We define two types of cross-features, namely model-variant cross-features and data-variant cross-features. 
Note that we use \emph{features} as synonymous to the \emph{last hidden layer activations}.
Model-variant cross-features are the features obtained from the received neighbors’ model with respect to the local dataset. 
These cross-features are computed locally at each agent after receiving the neighbors’ model parameters. 
Communicating the neighbors’ model parameters is a necessary step in any gossip-based decentralized algorithm \cite{d-psgd}. 
Data-variant cross-features are the features obtained from the local model with respect to its neighbors’ datasets. These cross-features are obtained through an additional round of communication. 

The $\mathcal{L}_{mv}$ loss term minimizes the $L_2$ distance between the local features and model-variant cross-features of the local data at each agent. Whereas $\mathcal{L}_{dv}$ minimizes the $L_2$ distance between the local- and data variant cross-feature representations of the same class. Figure.~\ref{fig:overview} provides an illustration of the loss terms introduced by the proposed framework. 
We validate the performance of the proposed framework through an exhaustive set of experiments on various vision datasets, model architectures, and graph topologies. We show that the proposed framework achieves superior performance on heterogeneous data compared to the current state-of-the-art method. We also report the communication and compute overheads required for proposed \emph{CCL} as compared to D-PSGD.

\textbf{Contributions:}
In summary, we make the following contributions.
\begin{itemize}
    \item We present a novel data-free knowledge distillation-based loss called \emph{Cross-feature Contrastive Loss (CCL)} for decentralized machine learning on heterogeneous data. 
    \item Through an exhaustive set of experiments, we show the advantages of our framework against the current state-of-the-art methods.
    \item We also report the communication and compute overheads incurred by the proposed framework.
\end{itemize}

\section{Background}
\label{sec:background}
\begin{figure}[ht]
  \centering
   \includegraphics[width=1.0\linewidth]{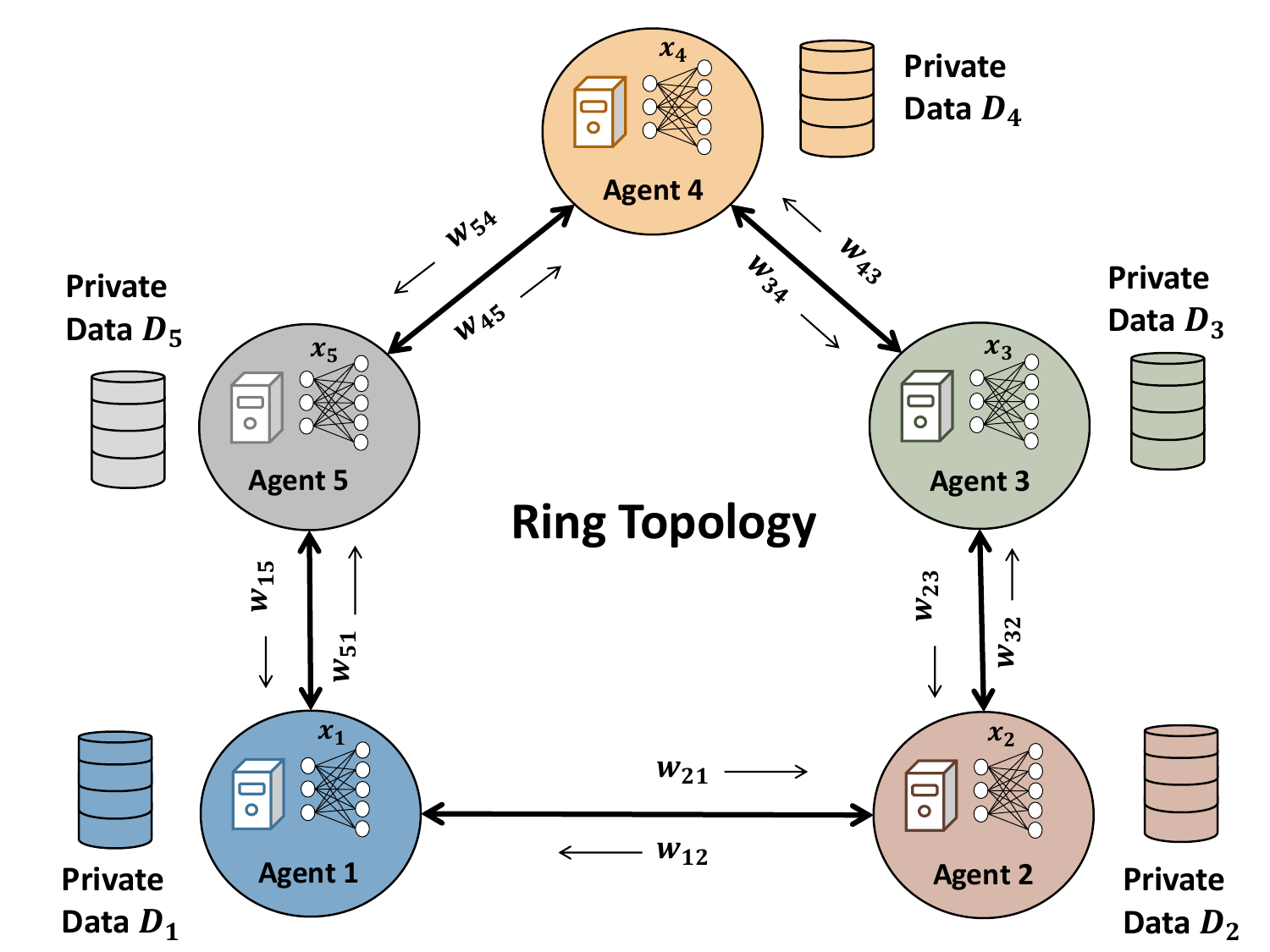}
   \caption{Decentralized training setup with 5 agents connected in a ring topology. Each agent has its own private dataset and a local model.}
   \label{fig:dec_overview}
\end{figure}

In this section, we provide the background on decentralized learning algorithms with peer-to-peer connections.
Figure.~\ref{fig:dec_overview} illustrates a decentralized setup with 5 agents connected in a ring topology.

The main goal of decentralized machine learning is to learn a global model using the knowledge extracted from the locally generated and stored data samples across $n$ edge devices/agents while maintaining privacy constraints. In particular, we solve the optimization problem of minimizing global loss function $f(x)$ distributed across $n$ agents as given in equation.~\ref{eq:1}. 
Note that $f_i$ is a local loss function (for example, cross-entropy loss $\mathcal{L}_{ce}$) defined in terms of the data sampled ($d_i$) from the local dataset $D_i$ at agent $i$ with model parameters $x_i$.
\begin{equation}
\label{eq:1}
\begin{split}
    \min \limits_{x \in \mathbb{R}^d} f(x) &= \frac{1}{n}\sum_{i=1}^n f_i(x), \\
    and \hspace{2mm} f_i(x) &= \mathbb{E}_{d_i \in D_i}[F_i(x; d_i)] \hspace{2mm} \forall i
\end{split}
\end{equation}
This is typically achieved by combining stochastic gradient descent \cite{sgd} with global consensus-based gossip averaging \cite{gossip}. 
The communication topology in this setup is modeled as a graph $G = ([n], E)$ with edges $\{i,j\} \in E$ if and only if agents $i$ and $j$ are connected by a communication link exchanging the messages directly. 
We represent $\mathcal{N}_i$ as the neighbors of $i$ including itself. It is assumed that the graph $G$ is strongly connected with self-loops i.e., there is a path from every agent to every other agent. 
The adjacency matrix of the graph $G$ is referred to as a mixing matrix $W$ where $w_{ij}$ is the weight associated with the edge $\{i,j\}$. Note that, weight $0$ indicates the absence of a direct edge between the agents. 
We assume that the mixing matrix is doubly stochastic and symmetric, similar to all previous works in decentralized learning \cite{d-psgd, qgm}. 
Further, the initial models and all the hyperparameters are synchronized at the beginning of the training. 
\begin{algorithm}[ht]
\textbf{Input:} Each agent $i \in [1,N]$ initializes model weights $x_i^{(0)}$, step size $\eta$, momentum coefficient $\beta$, and  mixing matrix $W=[w_{ij}]_{i,j \in [1,N]}$.\\

Each agent simultaneously implements the 
T\text{\scriptsize RAIN}( ) procedure\\
1.  \textbf{procedure} T\text{\scriptsize RAIN}( ) \\
2.  \hspace{4mm}\textbf{for} k=$0,1,\hdots,K-1$ \textbf{do}\\
3.  \hspace*{8mm}$d_i^{k} \sim D_i$\\
4.  \hspace*{8mm}$g_{i}^{k}=\nabla_x F_i(d_i^{k}; x_i^{k}) $ \\
5. \hspace*{8mm}$m_i^{k}= \beta m_i^{(k-1)} + g_i^{k}$\\
6.  \hspace*{8mm}$x_{i}^{k+\frac{1}{2}}=x_i^{k}-\eta m_i^{k}$\\
7.  \hspace*{8mm}S\text{\scriptsize END}R\text{\scriptsize ECEIVE}($x_{i}^{k+\frac{1}{2}}$)\\
8.  \hspace*{8mm}$x_i^{(k+1)}=\sum_{j\in \mathcal{N}_i} w_{ij} x_{j}^{k+\frac{1}{2}}$ \hfill \textcolor{gray!80}{// gossip averaging}\\
9.  \textbf{return}
\caption{Decentralized Learning with \textit{DSGDm} \cite{d-psgd}}
\label{alg:dl}
\end{algorithm}

Algorithm.~\ref{alg:dl} describes the flow of Decentralized Stochastic Gradient Descent with momentum (DSGDm). There are three main stages in any traditional decentralized learning method - (a) Local update, (b) Communication, and (c) Gossip averaging. At every iteration, each agent computes the gradients using local data and updates its model parameters as shown in line 6 of Alg.~\ref{alg:dl}. Then these updated model parameters are communicated to the neighbors as shown in line 7 of Alg.~\ref{alg:dl}. Finally, in the gossip averaging step, the local model parameters are averaged with the received model parameters of the neighbors using the mixing weights (shown in line 8 of Alg.~\ref{alg:dl}).
The convergence of the DSGDm algorithm assumes the data distribution across the agents to be Independent and Identically Distributed (IID).

\section{Related Work}

Decentralized Parallel Stochastic Gradient Descent (DSGD) \cite{d-psgd} is the first work to show that decentralized algorithms can converge at the same rate as their centralized counterparts \cite{dean2012large}. DSGD algorithm combines Stochastic Gradient Descent (SGD) with a gossip averaging algorithm \cite{gossip}. 
A momentum version of DSGD referred to as Decentralized Momentum Stochastic Gradient Descent (DSGDm) was proposed in \cite{balu2021decentralized}.
Further, Stochastic Gradient Push (SGP) \cite{sgp} extends the scope of DSGD to directed and time-varying graphs.
Recently, a unified framework for analyzing gossip-based decentralized SGD methods and the best-known convergence guarantees was presented in \cite{koloskova2020unified}. However, all of these above-mentioned algorithms assume the data distribution to be IID.

One of the core challenges in decentralized learning is tackling data that is not identically distributed among agents.
A wide range of algorithms were proposed in the literature to deal with the heterogeneous data.
The Methods such as Gradient Tracking \cite{gt}, and Momentum Tracking \cite{mt} track the global gradient and use it for the local update. This reduces variation in the local gradients across the agents and hence is more robust to heterogeneous data. 
Similarly, CGA\cite{CGA} and NGC \cite{ngc} also improve the performance by reducing the variation local gradient by utilizing the cross-gradient information. However, all these techniques incur $2\times$ communication overhead.
The authors in \cite{qgm} introduce Quasi-Global Momentum (QGM), a decentralized learning method that mimics the global synchronization of momentum buffer to mitigate the difficulties of decentralized learning on heterogeneous data.
Recently, RelaySGD was presented in \cite{relaysgd} that replaces the gossip averaging step with RelaySum \cite{zhang2019bp}. RelaySGD improves the performance of heterogeneous data by utilizing the delayed information in the RelaySum step. However, this technique only works on a spanning tree and the improvements do not scale well with the graph size.  

Knowledge distillation methods are well established for heterogeneous data in federated learning setups with a central server. However, there are only a few methods \cite{li2021decentralized, idkd} that explore knowledge distillation for decentralized learning on heterogeneous data. 
Decentralized federated learning via mutual knowledge
transfer (Def-KT) \cite{li2021decentralized} replaces gossip averaging with mutual knowledge transfer-based model fusion. In the Def-KT method, only a subset of agents are trained at a time while the other agents participate in model fusion. This deviates from the standard decentralized setup we use where all agents are trained parallely. 
In-Distribution Knowledge Distribution (IDKD) proposed in \cite{idkd} uses a public dataset and an Out-Of-Distribution detector to homogenize the data across decentralized agents. 
Orthogonal to these methods, we explore data-free knowledge distillation across agents through the proposed \textit{Cross-feature Contrastive Loss}. We compare the proposed \textit{CCL} method with QGM \cite{qgm} and RelaySGD \cite{relaysgd}, the current state-of-the-art methods in decentralized learning on heterogeneous data that do not incur any communication overhead or public dataset access.

\section{Cross-feature Contrastive Loss}

We propose the \textit{Cross-feature Contrastive Loss (CCL)} which aims to improve the performance of decentralized learning on heterogeneous data. \textit{CCL} introduces the concept of cross-features.

\begin{algorithm}[ht]
\textbf{Input:} Each agent $i \in [1,n]$ initializes model weights $x_i^{(0)}$, step size $\eta$, momentum coefficient $\beta$, averaging rate $\gamma$, contrastive loss coefficients $\lambda_m, \lambda_d$, mixing matrix $W=[w_{ij}]_{i,j \in [1,n]}$, number of classes C, $\mathcal{N}_i$ represents neighbors of $i$ including itself.\\

Each agent simultaneously implements 
T\text{\scriptsize RAIN}( ) procedure\\
1.   \textbf{procedure} T\text{\scriptsize RAIN}( ) \\
2.   \hspace{4mm}\textbf{for} k = $0,1,\hdots,K-1$ \textbf{do}\\
3.   \hspace*{8mm}S\text{\scriptsize END}R\text{\scriptsize ECEIVE}($x^i_{k}$)\\
4.   \hspace*{8mm}$d_i^{k} \sim D_i$\\
5.   \hspace*{8mm}\textbf{for} each neighbor $j \in \mathcal{N}_i$ \textbf{do}\\
6.   \hspace*{12mm}$z_{ji}^{k}=\phi(x_i^k; d_i^k)$\\
7.   \hspace*{12mm}\text{Compute} $\bar{z}_{ji}^{k}(c) \text{ i.e., the class-wise sum}$\\
8.   \hspace*{12mm}S\text{\scriptsize END}R\text{\scriptsize ECEIVE}($\{\bar{z}_{ji}^{k}(c), \text{count}(c)\}_{c=1}^C$)\\
9.    \hspace*{9mm}\textbf{end}\\
10. \hspace*{8mm}$\bar{z}^{k}(c)= \frac{1}{|c|}\sum_j \bar{z}_{ij}^{k}(c) \hspace{2mm} \forall \hspace{1mm} c \in [1,C]$\\
11. \hspace*{8mm}$CCL_i=\lambda_d\mathcal{L}_{dv}(z_{ii}^k, \bar{z}^k)+\lambda_m\mathcal{L}_{mv}(z_{ii}^k, \{z_{ji}\}_{\forall j})$ \\
12. \hspace*{8mm}$g_{i}^{k}=\nabla_x [\mathcal{L}_{ce}(x_i^k, d_i^k)+CCL_i] $ \\
13. \hspace*{8mm}$m_i^{k}= \beta \hat{m}_i^{(k-1)} + g_i^{k}$\\
14. \hspace*{8mm}$x_i^{(k+1)}=(\sum_{j\in \mathcal{N}_i} w_{ij} x_j^{k}) - \eta m_i^k$\\
15. \hspace*{8mm}$\hat{m}_i^{k}=\beta \hat{m}_i^{(k-1)}+(1-\beta)\frac{x_i^k - x_i^{(k+1)}}{\eta}$\\
16. \hspace{4mm}\textbf{end}\\
17. \textbf{return}
\caption{Decentralized Learning with \textit{CCL}}
\label{alg:CCL}
\end{algorithm}

\textbf{\emph{Cross-features ($z_{ij}$)}}: For an agent $i$ with model parameters $x_i$ connected to neighbor $j$ that has local dataset $D_j$, the cross-features are the last layer hidden representation obtained from the model parameters $x_i$, evaluated on mini-batch $d_j$ sampled from dataset $D_j$.
\begin{equation}
\label{eq:cf}
\begin{split}
    z_{ij}= \phi(x_i; d_j)
\end{split}
\end{equation}

Note that $\phi$ represents the neural network up to the last hidden layer (excluding the classifier) and all the definitions are provided with respect to an agent $i$. $z_{ii}$ represents the local feature representation i.e., $\phi(x_i, d_i)$. 
We define two types of cross-features, namely model-variant and data-variant cross-features.
\textit{Model-variant cross-features} ($\{z_{ji} | j \in \mathcal{N}_i \}$) are the features obtained from the received neighbors’ model $x_j$ with respect to the local dataset $d_i$. 
These cross-features are computed locally at each agent after receiving the neighbors’ model parameters. 
\textit{Data-variant cross-features} ($\{z_{ij} | j \in \mathcal{N}_i \}$) are the features obtained from the local model $x_i$ with respect to its neighbors’ datasets $d_j$. These cross-features are received through an additional communication round. 

Inspired by knowledge distillation methods, we introduce two different contrastive loss terms on cross-features. 
(a) \textbf{Model-variant contrastive loss} ($\mathcal{L}_{mv}$): At each agent $i$, $\mathcal{L}_{mv}$ minimizes the $L_2$ distance between the local feature representation $z_{ii}$ and the model-variant cross-features $z_{ji}$ for each data-point $q \in d_i$.
\begin{equation}
\label{eq:mvcl}
\begin{split}
    \mathcal{L}_{mv}(z_{ii}, \{z_{ji}\}_{\forall j}) &= \sum_{j \in \mathcal{N}_i} \frac{1}{|d_i|} \sum_{q \in d_i}||z_{ii}^q-z_{ji}^q||_2^2 \\
\end{split}
\end{equation}
The model-variant contrastive loss ensures that the model parameters on the neighboring agents are similar by enforcing the models to have similar representations for a given input sample. This reduces the variation in model parameters across the agents caused by the data heterogeneity. 

\noindent (b) \textbf{Data-variant contrastive loss} ($\mathcal{L}_{dv}$): To compute this loss, we first generate the neighborhood's representation $\bar{z}(c)$ for each class $c \in [1, C]$  using the data-variant cross-features $z_{ij}$'s as shown in Equation.~\ref{eq:dvcl}. 
Now at every agent $i$, $\mathcal{L}_{dv}$ minimizes the $L_2$ distance between the local representation $z_{ii}$ and the neighborhood's representation of the same class.
\begin{equation}
\label{eq:dvcl}
\begin{split}
    &\bar{z}(c) = \frac{1}{|c|}\sum_{j \in \mathcal{N}_i} \sum_{q \in d_j}z_{ij}^q\mathds{1}_c(z_{ij}^q) \hspace{2mm}\forall \hspace{1mm}c \in [1,C]\\
    &\mathcal{L}_{dv}(z_{ii}, \bar{z}) = \frac{1}{|d_i|} \sum_{q \in d_i} \sum_{c=1}^C||z_{ii}^q-\bar{z}(c)||_2^2 * \mathds{1}_c(z_{ii}^q) \\
\end{split}
\end{equation}
Here $|c|$ represents the total number of samples in the set $\{z_{ij}| j \in \mathcal{N}_i\}$ that belongs to class $c$ and $|d_i|$ is the mini-batch size. $\mathds{1}_c(z)$ is an indicator function that outputs 1 if $z$ belongs to class $c$. Since $\mathcal{L}_{dv}(z_{ii}, \bar{z})$ only uses the averaged representation of data-variant cross-features for each class, we sum these cross-features class-wise and communicate this sum along with the class count to the respective neighbors.
The data-variant contrastive loss ensures that the feature representations of a particular class are similar across the agents reducing the disparities caused due to data heterogeneity. 

Algorithm.~\ref{alg:CCL} presents the decentralized learning algorithm with the proposed \textit{CCL}. Each agent minimizes the contrastive loss terms along with the traditional cross-entropy loss as shown in Equation.~\ref{eq:loss}. $\lambda_m$ and $\lambda_d$ are the hyper-parameters that weigh the model-variant and data-variant contrastive loss respectively. 
\begin{equation}
\label{eq:loss}
\begin{split}
    \mathcal{L}_i = \mathcal{L}_{ce}(x_i,d_i)+\lambda_m\mathcal{L}_{mv}(z_{ii}, \{z_{ji}\}_{\forall j}) +\lambda_d\mathcal{L}_{dv}(z_{ii}, \bar{z})  \\
\end{split}
\end{equation}
It has been shown in \cite{qgm} that quasi-global momentum works better than local momentum for decentralized learning with heterogeneous data. Hence, we employ quasi-global momentum with the proposed loss as described in Algorithm.~\ref{alg:CCL}.

\section{Experiments}
\label{sec:results}

In this section, we analyze the performance of the proposed Cross-feature Contrastive Loss and compare it with the baselines DSGDm-N \cite{d-psgd}, RelaySGD \cite{relaysgd}, and the current state-of-the-art QG-DSGDm-N \cite{qgm}. The source code is available at \url{https://github.com/aparna-aketi/Cross_feature_Contrastive_Loss}

\subsection{Experimental Setup}
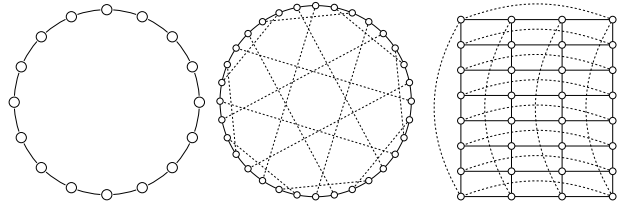
\begin{figure}[ht]
	\hfill
	\resizebox{1.0\linewidth}{!}{
		\resizebox{.28\linewidth}{!}{
			\begin{tikzpicture}[scale=1]
			\def \n {16}
			\def \radius {3.0cm}
			\def \margin {-4} %
			
			\foreach \s in {1,...,\n}
			{
				\node[draw, circle] at ({-360/\n * (\s - 1) + 90}:\radius) {};
				\draw[-, >=latex] ({-360/\n * (\s - 1)+\margin + 90}:\radius)
				arc ({-360/\n * (\s - 1)+\margin + 90}:{-360/\n * (\s)-\margin + 90}:\radius);
			}
			\end{tikzpicture}
   
		}
		\hfill
		\resizebox{.28\linewidth}{!}{
			\begin{tikzpicture}[scale=1]
			\def \n {32}
			\def \radius {5.0cm}
			\def \margin {-2} %
			
			\foreach \s in {1,...,\n}
			{
				\node[draw, circle] (\s) at ({-360/\n * (\s - 1) + 90}:\radius) {};
				\draw[-, >=latex] ({-360/\n * (\s - 1)+\margin + 90}:\radius)
				arc ({-360/\n * (\s - 1)+\margin + 90}:{-360/\n * (\s)-\margin + 90}:\radius);
			}
            \path [dashed, -] (1) edge node[left] {} (20);
            \path [dashed, -] (4) edge node[left] {} (17);
            \path [dashed, -] (9) edge node[left] {} (28);
            \path [dashed, -] (12) edge node[left] {} (25);
            \path [dashed, -] (5) edge node[left] {} (24);
            \path [dashed, -] (8) edge node[left] {} (21);
            \path [dashed, -] (13) edge node[left] {} (32);
            \path [dashed, -] (16) edge node[left] {} (29);

            \path [dashed, -] (2) edge node[left] {} (7);
            \path [dashed, -] (6) edge node[left] {} (11);
            \path [dashed, -] (10) edge node[left] {} (15);
            \path [dashed, -] (14) edge node[left] {} (19);
            \path [dashed, -] (18) edge node[left] {} (23);
            \path [dashed, -] (22) edge node[left] {} (27);
            \path [dashed, -] (26) edge node[left] {} (31);
            \path [dashed, -] (30) edge node[left] {} (3);
			\end{tikzpicture}
		}
		\hfill
		\resizebox{.27\linewidth}{!}{
			\begin{tikzpicture}[scale=2.5]
            \node[shape=circle,draw=black] (1) at (-1,-1.5) {};
            \node[shape=circle,draw=black] (2) at (0,-1.5) {};
			\node[shape=circle,draw=black] (3) at (1, -1.5) {};
			\node[shape=circle,draw=black] (4) at (2,-1.5) {};
            \node[shape=circle,draw=black] (5) at (-1,-1.0) {};
            \node[shape=circle,draw=black] (6) at (0,-1.0) {};
			\node[shape=circle,draw=black] (7) at (1, -1.0) {};
			\node[shape=circle,draw=black] (8) at (2,-1.0) {};
            \node[shape=circle,draw=black] (9) at (-1,-0.5) {};
            \node[shape=circle,draw=black] (10) at (0,-0.5) {};
			\node[shape=circle,draw=black] (11) at (1, -0.5) {};
			\node[shape=circle,draw=black] (12) at (2,-0.5) {};
			\node[shape=circle,draw=black] (13) at (-1,0) {};
			\node[shape=circle,draw=black] (14) at (0,0) {};
			\node[shape=circle,draw=black] (15) at (1, 0) {};
			\node[shape=circle,draw=black] (16) at (2,0) {};

            \node[shape=circle,draw=black] (17) at (-1,0.5) {};
			\node[shape=circle,draw=black] (18) at (0,0.5) {};
			\node[shape=circle,draw=black] (19) at (1,0.5) {};
			\node[shape=circle,draw=black] (20) at (2,0.5) {};
            \node[shape=circle,draw=black] (21) at (-1,1) {};
			\node[shape=circle,draw=black] (22) at (0,1) {};
			\node[shape=circle,draw=black] (23) at (1,1) {};
			\node[shape=circle,draw=black] (24) at (2,1) {};
			\node[shape=circle,draw=black] (25) at (-1,1.5) {};
			\node[shape=circle,draw=black] (26) at (0,1.5) {};
			\node[shape=circle,draw=black] (27) at (1, 1.5) {};
			\node[shape=circle,draw=black] (28) at (2,1.5) {};
			\node[shape=circle,draw=black] (29) at (-1,2) {};
            \node[shape=circle,draw=black] (30) at (0,2) {};
			\node[shape=circle,draw=black] (31) at (1, 2) {};
			\node[shape=circle,draw=black] (32) at (2,2) {};
            
			every edge/.style={draw=black}]
            \path [dashed,-] (1) edge[bend left=30] node[left] {} (29);
            \path [dashed,-] (2) edge[bend left=30] node[left] {} (30);
             \path [dashed,-] (3) edge[bend left=30] node[left] {} (31);
             \path [dashed,-] (4) edge[bend left=30] node[left] {} (32);
            
			\path [-] (1) edge node[left] {} (2);
			\path [-] (2) edge node[left] {} (3);
            \path [-] (3) edge node[left] {} (4);
            \path [-] (1) edge node[left] {} (5);
            \path [-] (2) edge node[left] {} (6);
            \path [-] (3) edge node[left] {} (7);
            \path [-] (4) edge node[left] {} (8);
			\path [-] (5) edge node[left] {} (6);
			\path [-] (6) edge node[left] {} (7);
            \path [-] (7) edge node[left] {} (8);
			\path [-] (5) edge node[left] {} (9);
            \path [-] (6) edge node[left] {} (10);
			\path [-] (7) edge node[left] {} (11);
			\path [-] (8) edge node[left] {} (12);
			\path [-] (9) edge node[left] {} (10);
			\path [-] (10) edge node[left] {} (11);
            \path [-] (11) edge node[left] {} (12);
            \path [-] (9) edge node[left] {} (13);
            \path [-] (10) edge node[left] {} (14);
			\path [-] (11) edge node[left] {} (15);
			\path [-] (12) edge node[left] {} (16);
            \path [-] (13) edge node[left] {} (14);
			\path [-] (14) edge node[left] {} (15);
            \path [-] (15) edge node[left] {} (16);
            \path [-] (13) edge node[left] {} (17);
			\path [-] (14) edge node[left] {} (18);
            \path [-] (15) edge node[left] {} (19);
            \path [-] (16) edge node[left] {} (20);
            \path [-] (17) edge node[left] {} (18);
            \path [-] (18) edge node[left] {} (19);
			\path [-] (19) edge node[left] {} (20);
            \path [-] (17) edge node[left] {} (21);
            \path [-] (18) edge node[left] {} (22);
			\path [-] (19) edge node[left] {} (23);
            \path [-] (20) edge node[left] {} (24);
            \path [-] (21) edge node[left] {} (22);
            \path [-] (22) edge node[left] {} (23);
			\path [-] (23) edge node[left] {} (24);
            \path [-] (21) edge node[left] {} (25);
            \path [-] (22) edge node[left] {} (26);
			\path [-] (23) edge node[left] {} (27);
            \path [-] (24) edge node[left] {} (28);
            \path [-] (25) edge node[left] {} (26);
            \path [-] (26) edge node[left] {} (27);
			\path [-] (27) edge node[left] {} (28);
            \path [-] (25) edge node[left] {} (29);
            \path [-] (26) edge node[left] {} (30);
			\path [-] (27) edge node[left] {} (31);
            \path [-] (28) edge node[left] {} (32);
            \path [-] (29) edge node[left] {} (30);
			\path [-] (30) edge node[left] {} (31);
            \path [-] (31) edge node[left] {} (32);
            \path [dashed,-] (4) edge[bend right=20] node[left] {} (1);
            \path [dashed,-] (8) edge[bend right=20] node[left] {} (5);
            \path [dashed,-] (12) edge[bend right=20] node[left] {} (9);
            \path [dashed,-] (16) edge[bend right=20] node[left] {} (13);
            \path [dashed,-] (20) edge[bend right=20] node[left] {} (17);
            \path [dashed,-] (24) edge[bend right=20] node[left] {} (21);
            \path [dashed,-] (28) edge[bend right=20] node[left] {} (25);
            \path [dashed,-] (32) edge[bend right=20] node[left] {} (29);
			\end{tikzpicture}
		}
	}
	\hfill\null
	\caption{Ring (left), Dyck (center), and Torus (right). }\label{fig:topologies}
\end{figure}

The efficiency of the proposed algorithm is demonstrated through experiments on a diverse set of datasets, model architectures, graph topologies, and graph sizes. We present the analysis on -- 
(a) \textbf{Datasets:} CIFAR-10 \cite{cifar}, CIFAR-100 \cite{cifar}, Fashion MNIST \cite{fmnist}, Imagenette \cite{imagenette}, and ImageNet \cite{imagenet}.
(b) \textbf{Model architectures:} ResNet-20, ResNet-18 \cite{resnet}, LeNet-5 \cite{lenet} and, MobileNet-V2 \cite{mobilnetv2}. All the models except LeNet-5 use Evonorm \cite{evonorm} as the activation-normalization layer as it is shown to be better suited for decentralized learning on heterogeneous data \cite{qgm}. LeNet-5 has no normalization layers.
(c) \textbf{Graph topologies:} Ring graph with 2 peers per agent, Dyck graph with 3 peers per agent, and Torus graph with 4 peers per agent (refer Figure~\ref{fig:topologies}).
(d) \textbf{Number of agents:} 8 to 40 agents.

For the decentralized setup, we use an undirected ring, undirected Dyck graph, and undirected torus graph topologies with a uniform mixing matrix. The undirected ring topology for any graph size has 3 peers per agent including itself and each edge has a weight of $\frac{1}{3}$. The undirected Dyck topology with 32 agents has 4 peers per agent including itself and each edge has a weight of $\frac{1}{4}$. The undirected torus topology with 32 agents has 5 peers per agent including itself and each edge has a weight of $\frac{1}{5}$. RelaySGD baseline only works on the spanning trees. Therefore, for a fair comparison, we use an undirected chain topology (spanning tree of ring topology) for all the RelaySGD experiments. We use the Dirichlet distribution to generate disjoint non-IID data across the agents similar to \cite{qgm}. The partitioned data across the agents is fixed, non-overlapping, and never shuffled across agents during the training. The degree of heterogeneity is regulated by the value of $\alpha$ -- the smaller the $\alpha$, the larger the non-IIDness across the agents.

The initial learning rate is either set to 0.1 (CIFAR-10, CIFAR-100) or 0.01 (Fashion MNIST, Imagenette) and is decayed by a factor of 10 after $50\%$ and $75\%$ of the training. 
We use a weight decay of $1e^{-4}$ and a mini-batch size of 32 per agent in all the experiments.
We use the Nesterov version of the Quasi-Global Momentum with a momentum coefficient of 0.9.
The stopping criteria is a fixed number of epochs. The experiments on CIFAR-10 are run for 200 epochs, CIFAR-100 and Imagenette for 100 epochs, and Fashion MNIST for 50 epochs. 
Note, DSGDm-N indicates Decentralized Stochastic Gradient Descent with Nesterov momentum, QG-DSGDm-N and the proposed \textit{CCL} uses a DSGD optimizer with Nesterov version of Quasi-Global Momentum. DSGDm-N,  QG-DSGDm-N, and RelaySGD utilized the cross-entropy loss whereas our framework uses the proposed \textit{Cross-feature Contrastive Loss} along with the cross-entropy loss. We use grid search on the set $\{1, 0.1, 0.01, 0.001\}$ to obtain the hyper-parameters $\lambda_m, \lambda_d$ for \textit{CCL}. 
We report the test accuracy of the consensus model averaged over three randomly chosen seeds. 
All our experiments were conducted on a system with an NVIDIA A40 card with 4 GPUs.
A detailed description of the setup and hyperparameters for each experiment is presented in Appendix.~\ref{apx:setup}.

\subsection{Results}

We evaluate the effectiveness of the \textit{Cross-feature Contrastive Loss} through an exhaustive set of experiments. We show that the proposed loss terms improve the accuracy as compared to simple cross-entropy loss. 

\begin{table}[ht]
\caption{Test accuracy of different decentralized algorithms evaluated on CIFAR-10, distributed with different degrees of heterogeneity (non-IID) trained on ResNet-20 over ring topologies. The results are averaged across all agents over three seeds where the standard deviation is indicated. We also include the results of the IID baseline as DSGDm-N (IID) where the local data is randomly partitioned independent of $\alpha$.}
\label{tab:cf10}
\small
\begin{center}
\resizebox{\columnwidth}{!}{
\begin{tabular}{cccc}
\hline
\multirow{ 2}{*}{Agents ($n$)} &\multirow{ 2}{*}{Method} &  \multicolumn{2}{c}{ResNet-20}   \\
\cmidrule(lr){3-4}  
& & $\alpha=0.1$ & $\alpha=0.01$  \\
\hline
\multirow{4}{*}{$16$} & DSGDm-N (IID) & \multicolumn{2}{c}{$89.61 \pm 0.43$}   \\
  & DSGDm-N \cite{d-psgd} &  $80.60 \pm 0.50$ & $58.78 \pm 9.63$ \\
  & RelaySGD \cite{relaysgd} & $73.81 \pm 1.97$ & $34.33 \pm 2.42$\\
  & QG-DSGDm-N \cite{qgm} &  $ 85.95 \pm 1.64$ & $77.16 \pm 7.02$  \\
  & CCL (ours) & $\mathbf{86.63} \pm 0.52$ & $\mathbf{81.29} \pm 0.36$ \\
 \hline 
\multirow{3}{*}{$32$} & DSGDm-N (IID) & \multicolumn{2}{c}{$88.13 \pm 0.32$}\\
 & DSGDm-N \cite{d-psgd} &  $76.46 \pm 1.32$ & $53.08 \pm  5.12$  \\
 & RelaySGD \cite{relaysgd} & $72.22 \pm 2.58$ & $38.16 \pm 1.34$ \\
 & QG-DSGDm-N \cite{qgm} &  $84.91 \pm 0.56$ & $75.70 \pm 0.80$  \\
 & CCL (ours) & $ \mathbf{85.25} \pm 0.52$ & $\mathbf{77.60} \pm 1.58$ \\
\hline
\end{tabular}
}
\end{center}
\end{table}

Table.~\ref{tab:cf10} presents the average test accuracy for training ResNet-20 architecture on the CIFAR-10 dataset with varying degrees of heterogeneity over the ring topology of 16 and 32 agents. 
We observe that \textit{CCL} outperforms QG-DSGDm-N for all models, graph sizes, and degree of heterogeneity with a significant performance gain varying from $0.34-4.13\%$. We also notice that the proposed framework has less variation in accuracy across various initial weight initialization (seeds) compared to the baselines. 
In our settings, we find that Relay-SGD with local momentum performs worse than DSGDm-N and doesn't scale with graph size. Note that \textit{Cross-feature Contrastive Loss} is an orthogonal technique to RelaySGD and can be used in synergy with RelaySGD. 

\begin{table}[ht]
\caption{Test accuracy of CIFAR-10 dataset trained on ResNet-20 over various graph topologies with 32 agents.
}
\label{tab:graphs}
\small
\begin{center}
\begin{tabular}{cccc}
\hline
Graph & Method & $\alpha=0.1$ & $\alpha=0.01$\\
\hline
  \multirow{3}{*}{Dyck} & DSGDm-N (IID) & \multicolumn{2}{c}{$88.89 \pm 0.10$} \\
& QG-DSGDm-N & $86.20 \pm 0.38$& $78.18 \pm 4.01$ \\
 & CCL (ours) & $\mathbf{86.78} \pm 0.41$ & $ \mathbf{80.63} \pm 1.54$\\
\hline
 \multirow{3}{*}{Torus} & DSGDm-N (IID) & \multicolumn{2}{c}{$88.86 \pm 0.31$} \\
  & QG-DSGDm-N & $87.75 \pm 0.39$ & $81.74 \pm 0.87$\\
& CCL (ours) & $ \mathbf{88.14} \pm 0.36$ & $\mathbf{82.30} \pm 0.56$ \\
\hline
\end{tabular}
\end{center}
\end{table}

To demonstrate the scalability and generalizability of \textit{CCL}, We present the results on various graph topologies and datasets.
Firstly, we train the CIFAR-10 dataset on ResNet-20 over the Dyck and Torus graphs with 32 agents to illustrate the impact of connectivity on the proposed framework. We observe a $0.4-2.5\%$ performance gains with varying connectivity (or spectral gap) as seen in Table.~\ref{tab:graphs}. This shows that the gains from the proposed technique are more pronounced in graphs with less connectivity such as ring graphs. 
We then evaluate \textit{CCL} on various image datasets such as Fashion MNIST and Imagenette and on challenging datasets such as CIFAR-100 and ImageNet. The proposed \textit{CCL} outperforms QG-DSGDm-N by $0.2-2.4\%$ across various datasets as shown in Table.~\ref{tab:datasets}.
Further, Table.~\ref{tab:imagenet} shows that the proposed method can achieve an average improvement of $1.1\%$ on the large-scale ImageNet dataset.

\begin{table*}[ht]
\caption{Test accuracy of different decentralized algorithms evaluated on various datasets, distributed with different degrees of heterogeneity over 16 agents ring topology with 16 agents.
}
\label{tab:datasets}
\small
\begin{center}
\begin{tabular}{ccccccc}
\hline
 \multirow{2}{*}{Method} & \multicolumn{2}{c}{Fashion MNIST (LeNet-5)} & \multicolumn{2}{c}{CIFAR-100 (ResNet-20)} & \multicolumn{2}{c}{Imagenette (MobileNet-V2)}\\
 \cmidrule(lr){2-3}  \cmidrule(lr){4-5} \cmidrule(lr){6-7} 
 & $\alpha=0.1$ & $\alpha=0.01$ & $\alpha=0.1$ & $\alpha=0.01$ & $\alpha=0.1$ & $\alpha=0.01$\\
\hline
 DSGDm-N (IID) & \multicolumn{2}{c}{$90.95 \pm 0.09$} & \multicolumn{2}{c}{$59.72 \pm 1.00$} & \multicolumn{2}{c}{$74.17 \pm 0.83$} \\
 QG-DSGDm-N & $88.92 \pm 0.50$ & $87.15 \pm 0.64$ & $52.33 \pm 3.59$ & $44.12 \pm 6.85$ & $65.94 \pm 1.17$ & $51.47 \pm  2.67$\\
 CCL (ours) & $\mathbf{90.21} \pm 0.34$ & $\mathbf{87.42} \pm 0.78$ & $\mathbf{54.20} \pm 0.86$ & $\mathbf{46.49} \pm 4.19$ & $\mathbf{66.14} \pm 0.84$ & $\mathbf{52.87} \pm  5.15$\\
\hline
\end{tabular}
\end{center}
\end{table*}

\begin{table}[ht]
\caption{Test accuracy of ImageNet trained on ResNet-18 architecture over ring topology with 16 agents.
}
\label{tab:imagenet}
\small
\begin{center}
\begin{tabular}{cccc}
\hline
Graph & Method & $\alpha=1$ & $\alpha=0.1$\\
\hline
  \multirow{3}{*}{Ring} & DSGDm-N (IID) & \multicolumn{2}{c}{$65.62 \pm 0.03$} \\
& QG-DSGDm-N & $64.09 \pm 1.49$& $58.11 \pm 3.81$ \\
 & CCL (ours) & $\mathbf{64.64} \pm 1.09$ & $ \mathbf{59.82} \pm 1.75$\\
\hline
\end{tabular}
\end{center}
\end{table}

\begin{figure}[ht]
  \centering
   \includegraphics[width=1.0\linewidth]{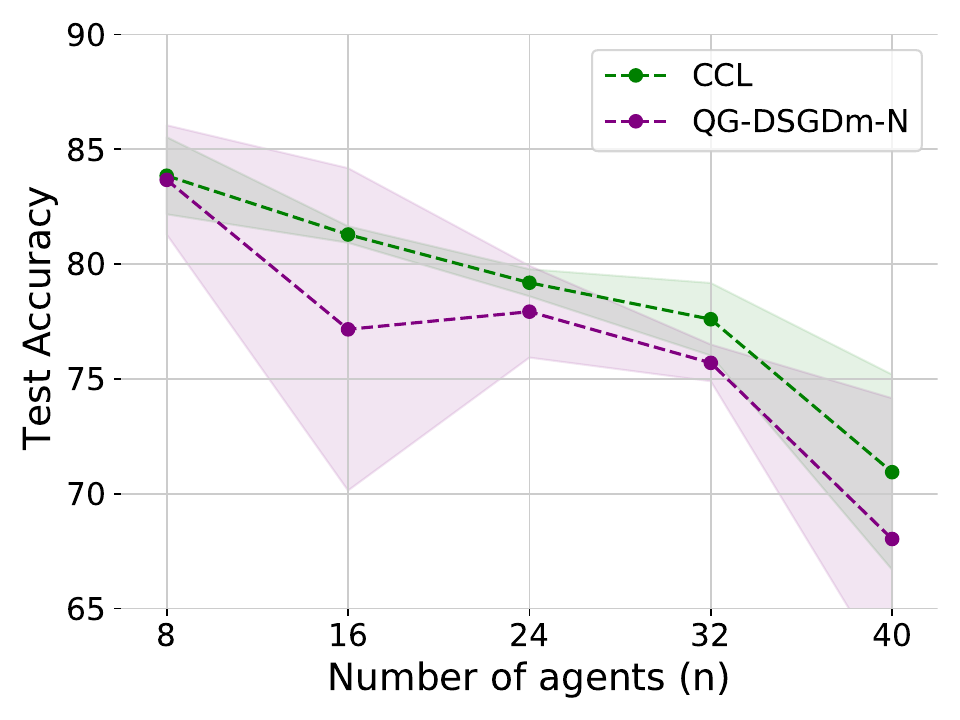}
   \caption{Test accuracy for the CIFAR-10 dataset trained on ResNet-20 architecture over varying sizes of ring topology with a skew of $\alpha=0.01$.}
   \label{fig:agents}
\end{figure} 

We then evaluate the scalability in decentralized settings by training CIFAR-10 on varying the size of the ring topology between 8 and 40 as shown in Figure.~\ref{fig:agents}. We observe that the proposed \textit{CCL} framework consistently outperforms the QG-DSGDm-N baseline over varying graph sizes by an average improvement of $2\%$.
In summary, the proposed \textit{Cross-feature Contrastive Loss} makes the decentralized training more robust to heterogeneity in the data distribution and has superior performance to all the comparison methods with an average improvement of $1.3\%$. Additional results are presented in Appendix.~\ref{apx:add_results}.

\begin{figure*}[ht]
\centering
\begin{subfigure}{0.32\textwidth}
    \includegraphics[width=\textwidth]{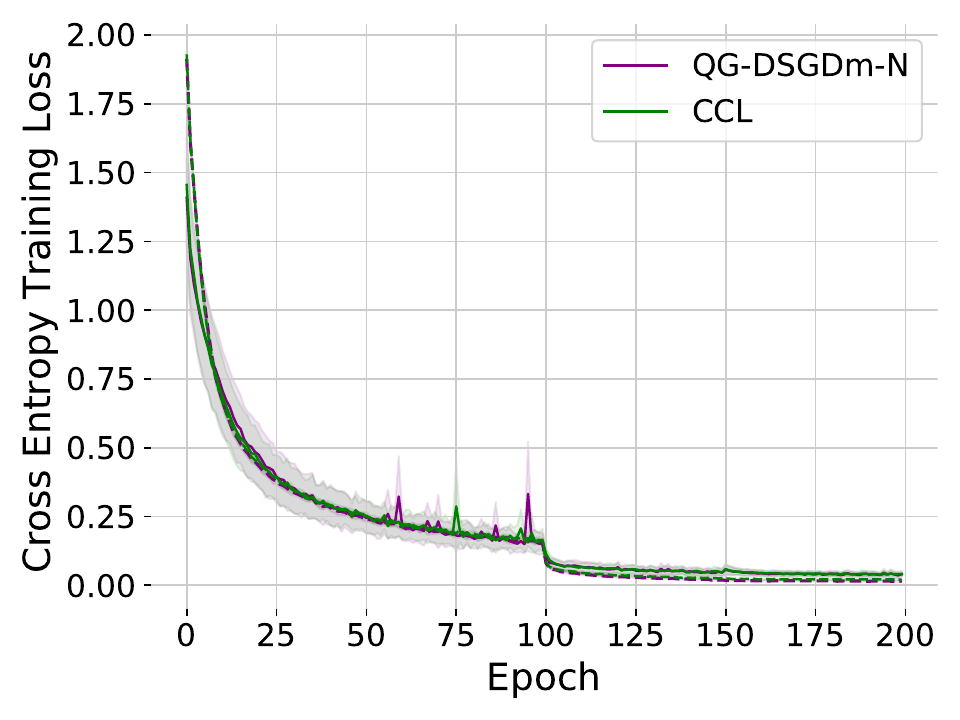}
    \caption{Cross-Entropy Training Loss ($\mathcal{L}_{ce}$)}
    \label{fig:lce_train}
\end{subfigure}
\hfill
\begin{subfigure}{0.32\textwidth}
    \includegraphics[width=\textwidth]{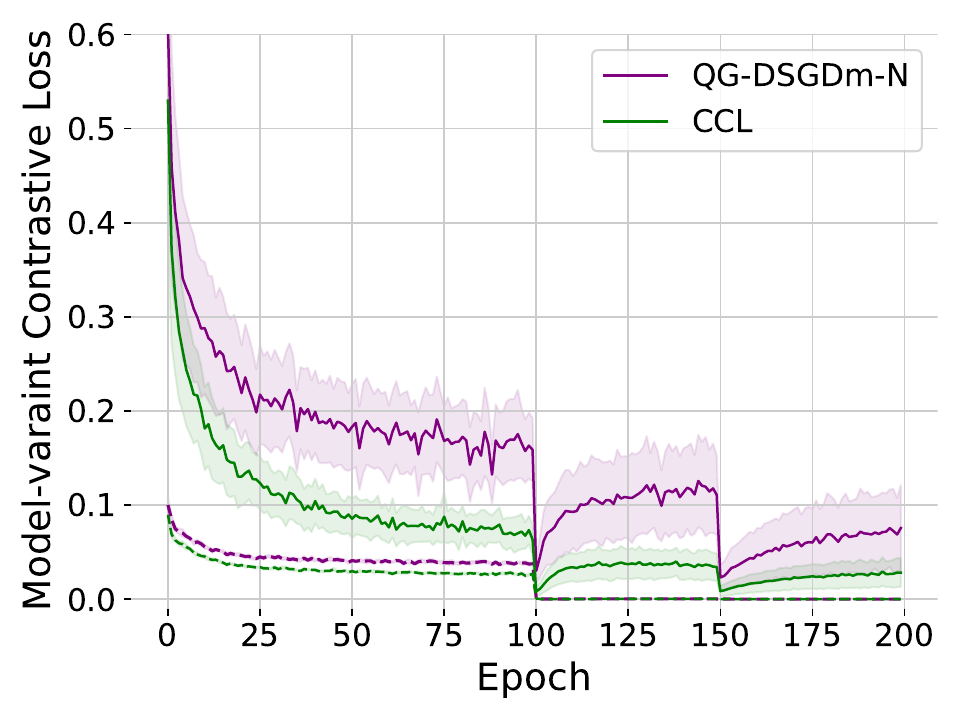}
    \caption{Model-Variant Contrastive Loss ($\mathcal{L}_{mv}$)}
    \label{fig:lmv_train}
\end{subfigure}
\hfill
\begin{subfigure}{0.32\textwidth}
    \includegraphics[width=\textwidth]{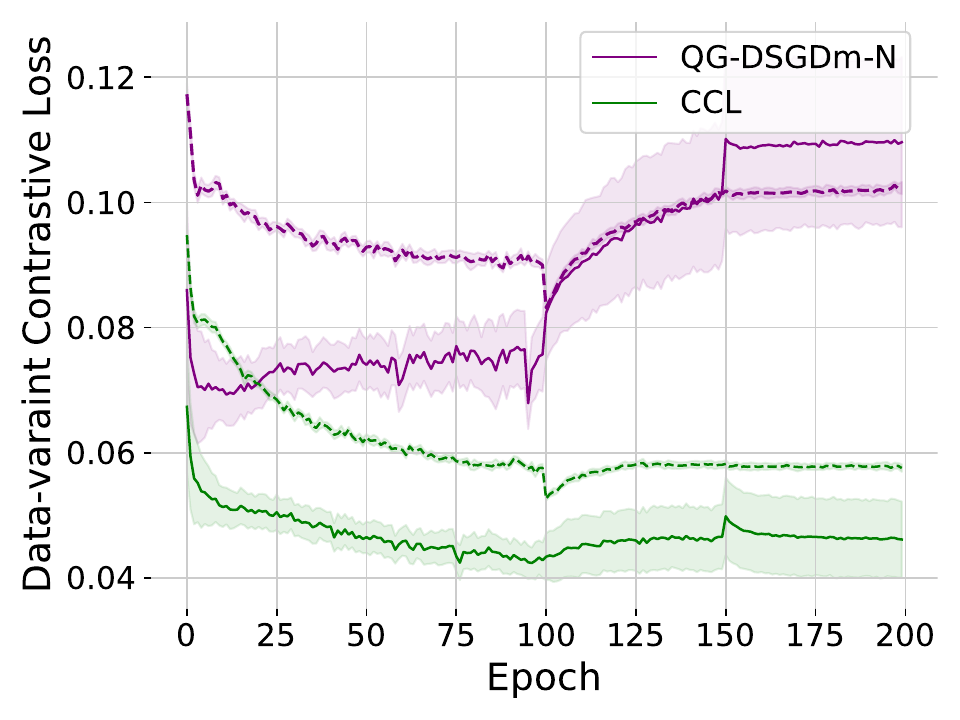}
    \caption{Data-Variant Contrastive Loss ($\mathcal{L}_{dv}$)}
    \label{fig:ldv_train}
\end{subfigure}        
\caption{Comparing various training loss terms for IID (dashed lines) and non-IID (solid lines) partitions of CIFAR-10 trained on ResNet-20 over a ring topology of 16 agents. We use $\alpha=10$ for IID data and $\alpha=0.1$ for non-IID data.
}
\label{fig:loss}
\end{figure*}

\subsection{Analysis}
In this section, we analyze the various aspects of the proposed \textit{CCL} terms such as the effect of the contrastive loss on IID vs. non-IID data, the choice of the loss function, and the contribution of each loss term. 

The proposed \textit{CCL} framework minimizes three different loss functions namely $\mathcal{L}_{ce}$, $\mathcal{L}_{mv}$, and $\mathcal{L}_{dv}$ where as the baseline methods (DSGDm-N, QG-DGSDm-N) only focus on $\mathcal{L}_{ce}$. We hypothesize that the cross-entropy loss $\mathcal{L}_{ce}$ alone does not capture the data heterogeneity across the agents. To address this we add two different contrastive loss terms explicitly representing the data skew. Figure.~\ref{fig:loss} measures the different training losses for both IID and non-IID distribution of the CIFAR-10 dataset. We observe that the training cross-entropy loss (Fig.~\ref{fig:lce_train}) for IID and non-IID data converges to zero even though there is a huge gap in the validation loss. 
However, Fig.~\ref{fig:lmv_train} shows that the model-variation contrastive loss for the baseline is much higher in non-IID settings compared to IID and hence is a good measure of data-heterogeneity. 
On the other hand, data-variant contrastive loss measures the variation in class representations across agents. Fig.~\ref{fig:ldv_train} shows that this variation is relatively stable throughout the training process for QG-DSGDm-N with IID Data. However, for QG-DSGDm-N with the non-IID setting, a significant increase in the variation of class representations across agents is evident. 
Note that the absolute value of the data-variant contrastive loss for the non-IID setting is lower than the IID setting because of limited shared class samples among neighboring agents (say 6 out of 32 in a mini-batch). 
The proposed \textit{CCL} framework explicitly minimizes the model-variant and data-variant contrastive loss. Fig.~\ref{fig:lmv_train}, \ref{fig:ldv_train} show that the proposed framework reduces the model variance and stabilizes the variance in class representations across agents resulting in better performance on heterogeneous data. 

\begin{table}[ht]
\caption{Test accuracy of CIFAR-10 dataset trained with different contrastive loss functions on ResNet-20 architecture over a ring topology.
}
\label{tab:loss}
\small
\begin{center}
\begin{tabular}{cccc}
\hline
 Loss function & Agents & $\alpha=0.1$ & $\alpha=0.01$\\
\hline
L1 Loss & \multirow{3}{*}{$16$} & $85.76 \pm 1.74$& $80.43 \pm 2.70$ \\
 MSE Loss & & $\mathbf{86.16} \pm 0.67$ & $81.29 \pm 0.36$\\
Cosine Loss & & $86.02 \pm 0.78$ & $\mathbf{82.36} \pm 0.93$\\
\hline
L1 Loss & \multirow{3}{*}{$32$} & $\mathbf{85.76} \pm 0.32$ & $76.13 \pm 2.59$\\
 MSE Loss& & $85.25 \pm 0.52$ & $\mathbf{77.60} \pm 1.58$ \\
Cosine Loss& & $85.71 \pm 0.27$ & $75.71 \pm 3.73$\\
\hline
\end{tabular}
\end{center}
\vspace{-2mm}
\end{table}

Cross-feature Contrastive Loss reduces the similarity between local feature representations and cross-feature representations. The similarity between the two representations can be determined using various measures. We explore three different similarity measures (or the loss functions) for the \textit{CCL} namely L1 loss, Mean Square Error (MSE) loss, and Cosine loss during the training phase. L1 loss measures the $L_1$ distance between the local and cross-features, MSE loss measures the $L_2$ distance, and Cosine loss measures the cosine distance. We observe that on average MSE loss provides better improvements as shown in Table.~\ref{tab:loss}. 

\begin{table}[ht]
\caption{Test accuracy of CIFAR-10 dataset trained with different components of contrastive loss on ResNet-20 over a ring topology.}
\label{tab:ccl_analysis}
\small
\begin{center}
\begin{tabular}{ccccc}
\hline
 $\mathcal{L}_{ce}$ & $\mathcal{L}_{mv}$ & $\mathcal{L}_{dv}$ & $\alpha=0.1$ & $\alpha=0.01$\\
\hline
\checkmark & \text{\sffamily x} & \text{\sffamily x} & $85.95 \pm 1.64$& $77.16 \pm 7.02$ \\
\checkmark  & \checkmark  & \text{\sffamily x} & $\mathbf{86.63} \pm 0.52$ & $80.55 \pm 1.61$\\
\checkmark  & \text{\sffamily x} & \checkmark  & $85.67 \pm 1.58$ & $77.78 \pm 4.01$\\
\checkmark  &\checkmark  & \checkmark  & $86.16 \pm 0.67$ & $\mathbf{81.29} \pm 0.36$\\ 
\hline
\end{tabular}
\end{center}
\end{table}

Cross-feature Contrastive Loss introduces two loss terms namely model-variant contrastive loss and data-variant contrastive loss. We evaluate the contribution of each of these loss terms to the improved accuracy in Table.~\ref{tab:ccl_analysis}. For lower skew ($\alpha=0.1$), we observe that the accuracy improvements can be mostly attributed to the addition of model-variant contrastive loss ($\mathcal{L}_{mv}$). Even in the case of higher skew ($\alpha=0.01$), the improvement can be majorly attributed to the model-variant contrastive loss. However, the maximum average test accuracy is obtained by adding both data-variant and model-variant contrastive loss terms.

\section{Discussion and Limitations}

The proposed \textit{Cross-feature Contrastive Loss} has two potential limitations -- (a) Compute overhead of the model-variant cross-features and (b) Communication overhead of the data-variant cross-features. 
Each agent has to compute model-variant cross-features at every iteration. This requires every agent to perform $p$ additional forward passes where $p$ is the number of peers/neighbors per agent. Assume that $c_f$ is the compute required for the forward pass. Now, the compute overhead can be given as $\mathcal{O}(p*c_f)$. Quantitatively, we measure the compute overhead as the fraction of additional compute required for the model-variant cross-features computation (Equation.~\ref{eq:compute}). 
\begin{equation}
\label{eq:compute}
\begin{split}
   \text{Compute overhead} = \frac{\text{Compute for cross-features}}{\text{Total compute}}
\end{split}
\end{equation}
Table.~\ref{tab:compute} presents the compute overhead for various settings. We observe that for a ring topology, the compute overhead is around $35-40\%$. This overhead shoots up to $57\%$ for a torus graph. Note that the compute overhead depends on the number of peers per agent rather than the total graph size. 

\begin{table}[ht]
\caption{Compute overhead incurred per agent due to \textit{Cross-feature Contrastive Loss}.
}
\label{tab:compute}
\small
\begin{center}
\begin{tabular}{cccc}
\hline
 \multirow{2}{*}{Dataset} & \multirow{2}{*}{Model} & \multirow{2}{*}{Peers} & Compute \\
 &  &  & overhead\\
\hline
Fashion-MNIST & LeNet-5 & 2 & $0.351$\\
CIFAR-10 & ResNet-20 & 2 & $0.397$\\
CIFAR-10 & ResNet-20 & 3 & $0.496$\\
CIFAR-10 & ResNet-20 & 4 & $0.567$\\
CIFAR-100 & ResNet-20 & 2 & $0.397$\\
ImageNette & MobileNet-V2 & 2 & $0.397$\\
\hline
\end{tabular}
\end{center}
\end{table}

Every agent communicates the class-wise sum of data-variant cross-features and class count along with the model parameters to each of their neighbors. The overhead is from the communication of data-variant cross-features. For example, for a dataset with $C$ classes and $r$ is the size of a cross-feature, every agent communicates $C$ cross-features of size $r$ (one for each class) and a vector of size $C$ indicating the number of samples per class in the mini-batch. 
Thus, the communication overhead can be given as $\mathcal{O}(p*C*(r+1))$. This overhead is negligible compared to the communication of model parameters. Table.~\ref{tab:commnication} compares the communication cost per iteration of \textit{CCL} with QG-DSGDm-N in MegaBytes (MB) for training various datasets over a 16-agents ring topology. We observe that the communication overhead is around $0.2\%$ for CIFAR-10, $2.3\%$ for CIFAR-100, $1.4\%$ for Fashion-MNIST, and $0.6\%$ for ImageNette. This shows that the communication overhead incurred by the proposed framework is insignificant. 
\begin{table}[ht]
\caption{Communication cost per agent per iteration over a ring graph of 16 agents.
}
\label{tab:commnication}
\small
\begin{center}
\resizebox{\columnwidth}{!}{
\begin{tabular}{llcl}
\hline
 \multirow{2}{*}{Dataset} & \multirow{2}{*}{Model} & \multirow{2}{*}{Method} & Comm. Cost  \\
 &  &  & (MB)\\
\hline
\multirow{2}{*}{Fashion-MNIST} & \multirow{2}{*}{LeNet-5} & QG-DSGDm-N  & $0.471$ ($1\times$)\\
 &  & \textit{CCL} & $0.477$  ($1.013\times$)\\
\hline
\multirow{2}{*}{CIFAR-10} & \multirow{2}{*}{ResNet-20} & QG-DSGDm-N  & $2.079$  ($1\times$)\\
 &  & \textit{CCL} & $2.084$  ($1.002\times$)\\
\hline
\multirow{2}{*}{CIFAR-100} & \multirow{2}{*}{ResNet-20} & QG-DSGDm-N  & $2.123$ ($1\times$) \\
 &  & \textit{CCL} & $2.173$ ($1.023\times$)\\
\hline
\multirow{2}{*}{ImageNette} & \multirow{2}{*}{MobileNet-V2} & QG-DSGDm-N  & $17.52$ ($1\times$)\\
 &  & \textit{CCL} & $17.62$  ($1.006\times$) \\
\hline
\end{tabular}
}
\vspace{-2mm}
\end{center}
\end{table}

Further, a minor limitation of the proposed \textit{CCL} is that it adds two additional hyper-parameters $\lambda_m, \lambda_d$. These hyperparameters need to be tuned similarly to the learning rate. We used a grid search to find these hyperparameters. 
We observed that these hyperparameters typically take a value of $0.1$ or $0.01$. 
We consider the exploration of the compute efficient \textit{CCL} and adaptive \textit{CCL} as potential future research directions.

\section{Conclusion}
Supporting decentralized training on heterogeneous data is a critical factor for machine learning applications to effectively harness the vast amounts of user-generated private data.
In this paper, we propose a novel \textit{Cross-feature Contrastive Loss} which is better suited for decentralized learning on heterogeneous data. In particular, minimizing proposed contrastive loss terms increases the similarity of local feature representations with the model-variant and data-variant cross-features. We evaluate the \textit{CCL} with Quasi-Global Momentum through an exhaustive set of experiments on various datasets, model architectures, and graph topologies. Our experiments confirm the superior performance of the \textit{Cross-feature Contrastive Loss}  compared to existing state-of-the-art methods for decentralized learning on heterogeneous data.

\section*{Acknowledgements}

This work was supported in part by, the Center for the Co-Design of Cognitive Systems (COCOSYS), a DARPA-sponsored JUMP center, the Semiconductor Research Corporation (SRC), and the National Science Foundation.

{\small
\bibliographystyle{ieee_fullname}
\bibliography{egbib}
}

\clearpage

\appendix
\section{Appendix}
\subsection{Experimental Setup Details}
\label{apx:setup}

For the decentralized setup, we use an undirected ring, undirected Dyck graph, and undirected torus graph topologies with a uniform mixing matrix. The undirected ring topology for any graph size has 3 peers per agent including itself and each edge has a weight of $\frac{1}{3}$. The undirected Dyck topology with 32 agents has 4 peers per agent including itself and each edge has a weight of $\frac{1}{4}$. The undirected torus topology with 32 agents has 5 peers per agent including itself and each edge has a weight of $\frac{1}{5}$.
All our experiments were conducted on a system with an NVIDIA A40 card with 4 GPUs. We report the test accuracy of the consensus model averaged over three randomly chosen seeds. The consensus model is obtained by averaging the model parameters across all agents using an all-reduce mechanism at the end of the training.

\subsubsection{Datasets}
In this section, we give a brief description of the datasets used in our experiments. We use a diverse set of datasets each originating from a different distribution of images to show the generalizability of the proposed techniques.

\textbf{CIFAR-10:} 
CIFAR-10 \cite{cifar} is an image classification dataset with 10 classes. The image samples are colored (3 input channels) and have a resolution of $32 \times 32$. 
There are $50,000$ training samples with $5000$ samples per class and $10,000$ test samples with $1000$ samples per class.

\textbf{CIFAR-100:} 
CIFAR-100 \cite{cifar} is an image classification dataset with 100 classes. The image samples are colored (3 input channels) and have a resolution of $32 \times 32$. There are $50,000$ training samples with $500$ samples per class and $10,000$ test samples with $100$ samples per class. CIFAR-100 classification is a harder task compared to CIFAR-10 as it has 100 classes with very few samples per class to learn from.

\textbf{Fashion MNIST:}
Fashion MNIST \cite{fmnist} is an image classification dataset with 10 classes. The image samples are in greyscale (1 input channel) and have a resolution of $28 \times 28$. There are $60,000$ training samples with $6000$ samples per class and $10,000$ test samples with $1000$ samples per class.

\textbf{Imagenette:}
Imagenette \cite{imagenette} is a 10-class subset of the ImageNet dataset. The image samples are colored (3 input channels) and have a resolution of $224 \times 224$. There are $9469$ training samples with roughly $950$ samples per class and $3925$ test samples. 

\textbf{ImageNet:}
ImageNet dataset \cite{imagenet} spans 1000 object classes and contains 1,281,167 training images, 50,000 validation images, and 100,000 test images. The image samples are colored (3 input channels) and have a resolution of $224 \times 224$.

\begin{figure*}[ht]
\centering
\begin{subfigure}{0.32\textwidth}
    \includegraphics[width=\textwidth]{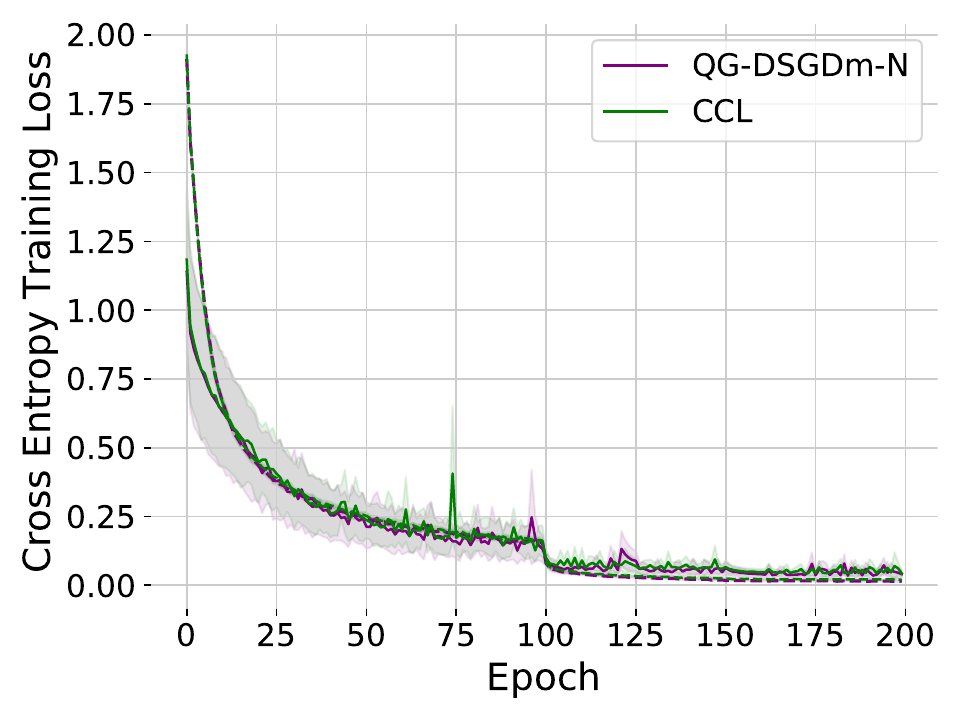}
    \caption{Cross-Entropy Training Loss ($\mathcal{L}_{ce}$)}
    \label{fig:lce_train_01}
\end{subfigure}
\hfill
\begin{subfigure}{0.32\textwidth}
    \includegraphics[width=\textwidth]{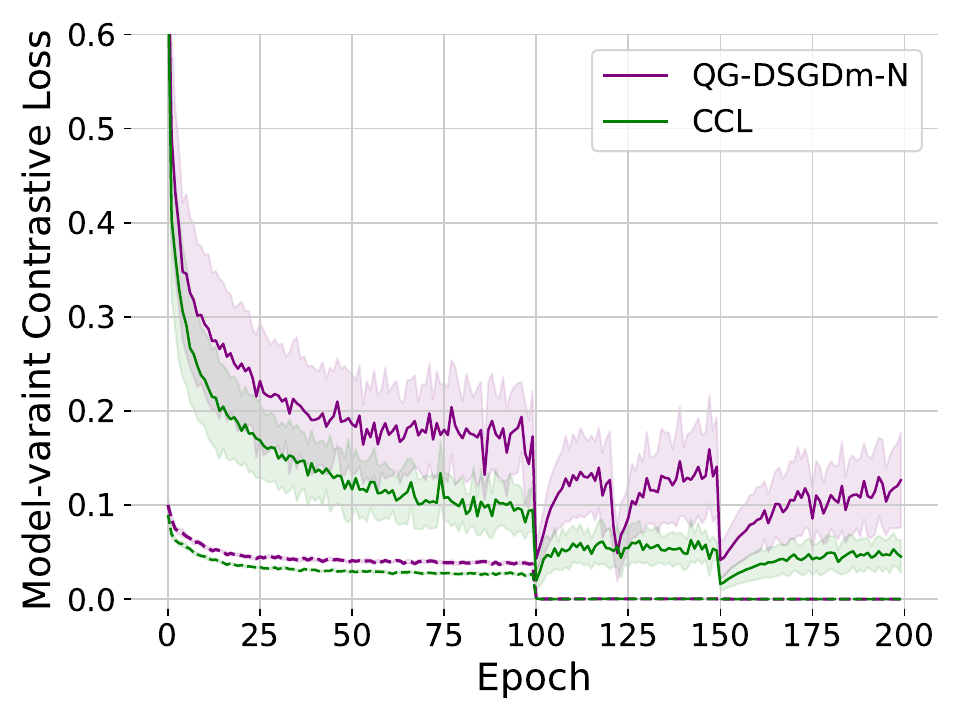}
    \caption{Model-Variant Contrastive Loss ($\mathcal{L}_{mv}$)}
    \label{fig:lmv_train_01}
\end{subfigure}
\hfill
\begin{subfigure}{0.32\textwidth}
    \includegraphics[width=\textwidth]{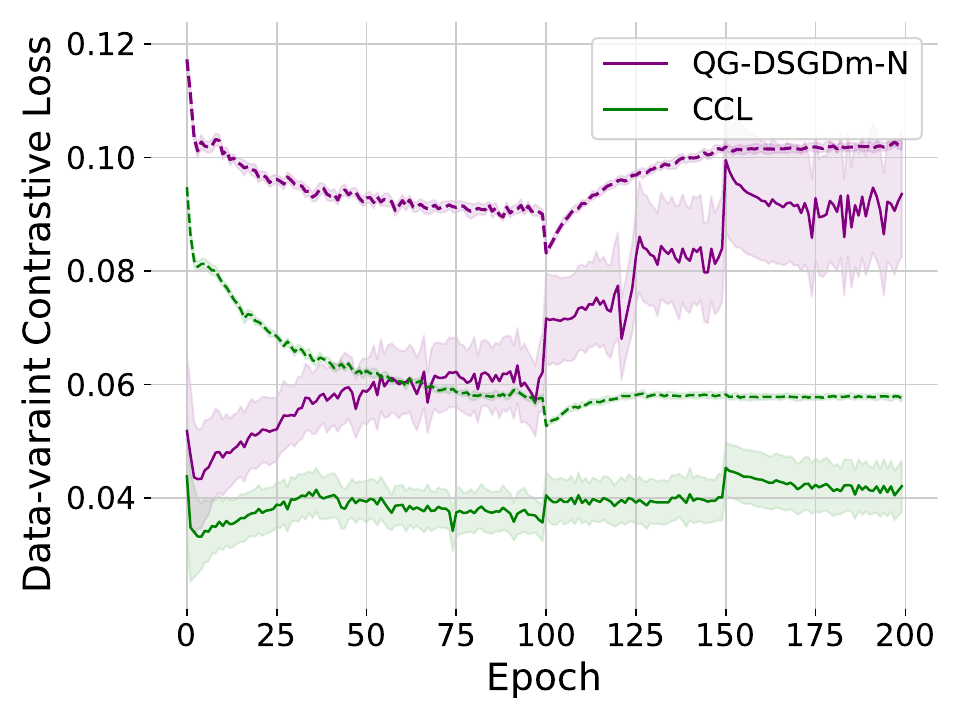}
    \caption{Data-Variant Contrastive Loss ($\mathcal{L}_{dv}$)}
    \label{fig:ldv_train_01}
\end{subfigure}        
\caption{Comparing various training loss terms for IID (dashed lines) and non-IID (solid lines) partitions of CIFAR-10 trained on ResNet-20 over a ring topology of 16 agents. We use $\alpha=10$ for IID data and $\alpha=0.01$ for non-IID data.
}
\label{fig:loss_01}
\end{figure*}

\subsubsection{Network Architecture}
We replace ReLU+BatchNorm layers of all the model architectures with EvoNorm-S0 as it was shown to be better suited for decentralized learning over non-IID distributions.

\textbf{ResNet-20:} For ResNet-20 \cite{resnet}, we use the standard architecture with $0.27M$ trainable parameters except that BatchNorm+ReLU layers are replaced by EvoNorm-S0.

\textbf{ResNet-18:} For ResNet-18 \cite{resnet}, we use the standard architecture with $11M$ trainable parameters except that BatchNorm+ReLU layers are replaced by EvoNorm-S0.

\textbf{LeNet-5:} For LeNet-5 \cite{lenet}, we use the standard architecture with $61,706$ trainable parameters.

\textbf{MobileNet-V2:} We use the the standard MobileNet-V2 \cite{mobilnetv2} architecture used for CIFAR dataset with $2.3M$ parameters except that BatchNorm+ReLU layers are replaced by EvoNorm-S0.

\subsubsection{Hyper-parameters}
This section presents a detailed description of the hyper-parameters used in our experiments. All the experiments were run for three randomly chosen seeds. We decay the step size by 10x after 50\% and 75\% of the training, unless mentioned otherwise.  For all the experiments, we have used a momentum of 0.9 with Nesterov, a weight decay of 0.0001, and a mini-batch size of 32 per agent. 
\begin{table}[ht]
\caption{The value of $\lambda_m, \lambda_v$ used for training CIFAR-10 with non-IID data using ResNet-20 architecture presented in Table 1}
\label{tab:cf10_hp}
\small
\begin{center}
\resizebox{\columnwidth}{!}{
\begin{tabular}{cccc}
\hline
\multirow{ 2}{*}{Agents ($n$)} &\multirow{ 2}{*}{Method} &  \multicolumn{2}{c}{ResNet-20}   \\
\cmidrule(lr){3-4}  
& & $\alpha=0.1$ & $\alpha=0.01$  \\
\hline
  16 & CCL (ours) & $0.01,0.0$ & $0.01,0.01$\\
 32 & CCL (ours) & $0.1,0.1$ & $0.1,0.1$ \\
\hline
\end{tabular}
}
\end{center}
\end{table}

\textbf{Hyper-parameters for experiments in Table 1:}
All the experiments have the stopping criteria set to 200 epochs. The initial learning rate is set to 0.1. 
We decay the step size by $10\times$ in multiple steps at $100^{th}$ and $150^{th}$ epoch. Table~\ref{tab:cf10_hp} presents values of the scaling factor $\lambda_m, \lambda_d$ used in the experiments. 

\begin{table}[ht]
\caption{The value of $\lambda_m, \lambda_v$ used for training various datasets with CCL (presented in Table 2).
}
\label{tab:datasets_hp}
\small
\begin{center}
\begin{tabular}{ccc}
\hline
Dataset & $\alpha=0.1$ & $\alpha=0.01$ \\
\hline
Fashion MNIST & $0.001, 0.001$ & $0.01, 0.01$\\
CIFAR-100 & $0.1, 0.1$ & $0.1, 0.1$ \\
Imagenette & $0.001, 0.001$ & $1.0, 1.0$\\
\hline
\end{tabular}
\end{center}
\end{table}

\textbf{Hyper-parameters for experiments in Table 2:}
All the experiments for CIFAR-100 and ImageNette have the stopping criteria set to 100 epochs and Fashion MNIST experiments have a stopping criteria of 50 epochs. The initial learning rate is set to 0.1 for CIFAR-100 and 0.01 for Fashion MNIST and Imagenette. 
Table~\ref{tab:datasets_hp} presents values of the scaling factor $\lambda_m, \lambda_d$ used in the experiments.  

\textbf{Hyper-parameters for experiments in Table 3:}
All the experiments have the stopping criteria set to 200 epochs. The initial learning rate is set to 0.1. 
We decay the step size by $10\times$ in multiple steps at $100^{th}$ and $150^{th}$ epoch. Table~\ref{tab:graphs_hp} presents values of the scaling factor $\lambda_m, \lambda_d$ used in the experiments. All the experiments on the Dyck and Torus graph use an averaging rate of 0.9 (instead of the default value of 1.0).

\begin{table}[ht]
\caption{The value of $\lambda_m, \lambda_v$ used for training CIFAR-10 datasets with CCL on ResNet-20 over various graph topologies (presented in Table 3).
}
\label{tab:graphs_hp}
\small
\begin{center}
\begin{tabular}{ccc}
\hline
Graph &  $\alpha=0.1$ & $\alpha=0.01$\\
\hline
  Dyck (32 agents) & $0.1, 0.1$ & $0.1, 0.1$ \\
 Torus (32 agents) & $0.1, 0.1$ &  $0.1, 0.1$ \\
\hline
\end{tabular}
\end{center}
\end{table}

\subsection{Additional Results}
\label{apx:add_results}
Figure.~\ref{fig:loss_01} measures the different training losses for both IID and non-IID distribution with $\alpha=0.01$ of the CIFAR-10 dataset trained on ResNet-20 architecture. We observe that the training cross-entropy loss (Fig.~\ref{fig:lce_train_01}) for IID and non-IID data converges to zero even though there is a huge gap in the validation loss.
However, Fig.~\ref{fig:lmv_train_01} shows that the model-variation contrastive loss for the baseline is much higher in non-IID settings compared to IID and hence is a good measure of data-heterogeneity. 
\begin{figure}[ht]
  \centering
   \includegraphics[width=1.0\linewidth]{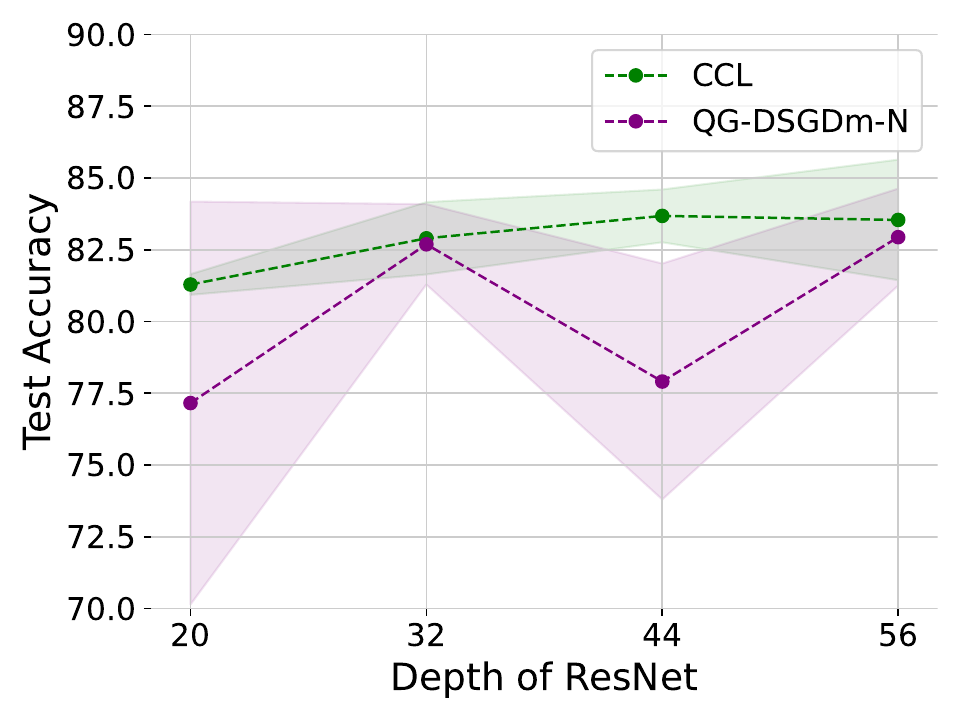}
   \caption{Test accuracy for the CIFAR-10 dataset trained on ResNet architecture with varying depth over 16-agent ring topology with a skew of $\alpha=0.01$.}
   \label{fig:depth}
\end{figure}
On the other hand, data-variant contrastive loss measures the variation in class representations across agents. Fig.~\ref{fig:ldv_train_01} shows that this variation is relatively stable throughout the training process for QG-DSGDm-N (baseline) with IID Data. 
However, for QG-DSGDm-N with the non-IID setting, a significant increase in the variation of class representations across agents is evident.  
The proposed \textit{CCL} framework explicitly minimizes the model-variant and data-variant contrastive loss. Fig.~\ref{fig:lmv_train_01} shows that the CCL helps in reducing the model variance compared to QG-DSGDm-N.  Fig.~\ref{fig:ldv_train_01} shows that CCL has a stable variation in class representations across agents compared to QG-DSGDm-N. This results in better performance of the proposed \textit{Cross-feature Contrastive Loss} for decentralized learning on heterogeneous data. 
Further, we evaluate the proposed \textit{CCL} on the varying depth of ResNet architecture with ring topology of 16 agents as shown in Figure.~\ref{fig:depth}. We observe that the proposed \textit{CCL} framework consistently outperforms the QG-DSGDm-N baseline over varying graph sizes by an average improvement of $2.68\%$.

\end{document}